# AN EMPIRICAL ASSESSMENT OF DEEP LEARNING APPROACHES TO TASK-ORIENTED DIALOG MANAGEMENT

ACCEPTED MANUSCRIPT

**Lukáš Matějů, David Griol, Zoraida Callejas, José Manuel Molina, Araceli Sanchis**
Technical University of Liberec, Faculty of Mechatronics, Informatics, and Interdisciplinary Studies,
Studentská 2, 46117, Liberec, Czech Republic
`lukas.mateju@tul.cz`
Software Engineering Department, University of Granada, Spain
`{dgriol, zoraida}@ugr.es`
Universidad Carlos III de Madrid, Computer Science Dept., Avda. de la Universidad 30, 28911 Leganés, Spain
`molina@ia.uc3m.es, masm@inf.uc3m.es`

# ABSTRACT

Deep learning is providing very positive results in areas related to conversational interfaces, such as speech recognition, but its potential benefit for dialog management has still not been fully studied. In this paper, we perform an assessment of different configurations for deep-learned dialog management with three dialog corpora from different application domains and varying in size, dimensionality and possible system responses. Our results have allowed us to identify several aspects that can have an impact on accuracy, including the approaches used for feature extraction, input representation, context consideration and the hyper-parameters of the deep neural networks employed.

# 1 Introduction

Conversational systems can be defined as computer programs that engage the user in a dialog that aims to be similar to that between humans [1, 2, 3, 4]. With advances in language technologies and Natural Language Processing (NLP), these interfaces have begun to play an increasingly important role in the design of human-machine interaction systems in a number of devices and intelligent environments [1] including online customer services that automatically reply to user questions about products or services reducing operation costs and enhancing user experience, entertainment conversational systems that chat with users in open domains, educational conversational systems that guide the users step by step and help them master the desired skills, personal assistants to interact with portable mobile terminal devices, intelligent quiz systems that assist users in decision-making purposes, smart home conversational assistants [5], and robots that offer spoken communication.

The increased number of applications that include conversational interfaces is supported by recent major advances in artificial intelligence (big data and deep learning to cite just two), language technologies (including automatic speech recognition and natural language processing), and device technologies (more powerful smartphones, use of sensors and context information, and increased connectivity).

One of the core aspects in the development of adaptive conversational interfaces is to design the Dialog Management (DM) strategy. This strategy defines the system's conversational behaviors in response to user utterances and environmental states. The design of this strategy is usually carried out in industry by handcrafting dialog strategies that are tightly coupled to the application domain in order to optimize the behavior of the conversational interface in that context.

This has motivated the research community to find ways for automating dialog learning by using statistical models trained with real conversations. Statistical approaches can model the variability in user behaviors and allow the exploration of a wide range of strategies. The application of machine learning approaches to DM strategy design is a rapidly growing research area [6, 7, 8, 9]. The main idea is to learn optimal strategies from corpora of real human-computer dialog data instead of relying on empirical design principles [10].

Statistical approaches to DM present additional important advantages. Rather than maintaining a single hypothesis for the dialog state, they maintain a distribution over many hypotheses for the correct dialog state. In addition, statistical methodologies choose actions using an optimization process, in which a developer specifies high-level goals and the optimization works out the detailed dialog plan. Finally, statistical DM systems have shown, in research settings, more robustness to speech recognition errors, yielding shorter dialogs with higher task completion rates [1].

In this paper, we evaluate the effect of different aspects related to the use of deep neural networks for the development of dialog managers on three dialog corpora of different sizes and complexity. Our proposal can be used to develop a statistical dialog manager that decides the current phase of the dialog by means of a classification process that considers the complete history of the dialog, which is one of the main advantages regarding the previously described statistical methodologies. Our results show the importance of the representation of the feature vectors employed as an input to the classifier, the ways in which context is encoded in them or/and in a multi-turn window, the effect of the configuration of the hyper-parameters of the network and the type of corpus employed.

The remainder of the paper is structured as follows. Section 2 presents the theoretical background of deep learning as well as summarizing current applications for conversational interfaces. In Section 3, the three corpora used within this work are introduced. Our experimental setup is described in Section 4 while the results are given in Section 5. The results achieved are discussed in detail in Section 6. Finally, the conclusions and guidelines for future work are presented in Section 7.

## 2 State of the art

As has been described in the previous section, different modules and processes must cooperate to achieve the main goal of a conversational interface. Automatic speech recognition (ASR) is the process of obtaining the text string corresponding to an acoustic input. Once the speech recognition component has recognized what the user uttered, it is necessary to understand what was said. Spoken language understanding (SLU) is the process of obtaining a semantic interpretation of a text string. Dialog management relies on the fundamental task of deciding what action or response a system should take in response to the user's input. There is no universally agreed definition of the tasks that this component has to carry out to make this decision. Traum et al. [11] state that the DM task involves four main tasks: (1) updating the dialog context; (2) providing a context for interpretation; (3) coordinating other modules; and (4) deciding the information to convey and when to do it.

Although DM is only one part of the information flow of a conversational interface, it can be seen as one of the most important tasks given that this component encapsulates the logic of the speech application. The selection of a particular action depends on multiple factors, such as the output of ASR (e.g., measures that define the reliability of the recognized information), the dialog interaction (e.g., the number of repairs carried out so far), the application domain (e.g., guidelines and regulations for customer service), and the responses and status of external back-ends, devices, and data repositories. Given that the actions of the system directly impact users, the dialog manager is largely responsible for user satisfaction. Because of these factors, the design of an appropriate DM strategy is at the core of the conversational interface engineering.

One of the simplest DM strategies is based on the definition of handcrafted rules. Although this approach has been deployed in many practical applications because of its simplicity, these applications only support a strict system-directed dialog interaction, in which at each turn the system directs the user by proposing a small number of choices for which there is a limited grammar or vocabulary to interpret the input.

The main trend of automating the DM task by means of machine learning approaches is increased use of data to improve the performance of conversational systems. Statistical models can be trained using corpora of human-computer dialogs with the goal of explicitly modeling the variability in user behavior that can be difficult to address by means of handwritten rules. The goal is to build systems that exhibit more robust performance, improved portability, better scalability, and easier adaptation to other tasks [12, 5, 9].

### 2.1 Statistical DM systems

Statistical approaches to DM can be classified into three main categories: example-based dialog management, dialog modeling based on reinforcement learning (RL), and corpus-based statistical dialog management. Example-based approaches can be considered a specific case of corpus-based statistical dialog management, given that they usually perform dialog modeling by means of prepared dialog examples [13, 14]. These approaches assume that the next system action can be predicted when the dialog manager finds dialog examples that have a similar dialog state to the current dialog state [15]. The best example is then selected from the candidate examples by calculating heuristic similarity measures between the current input and the example.

The most widespread methodology for dialog modeling based on reinforcement learning involves modeling human-computer interaction as an optimization problem using Markov decision processes (MDPs) and reinforcement learning methods [16, 17]. The main drawback of this approach is that the large state space required for representing all the possible dialog paths in practical spoken conversational interfaces makes its direct representation intractable. In addition, while exact solution algorithms do exist, they do not scale to problems with more than a few states/actions [18].

Partially observable MDPs (POMDPs) outperform MDP-based dialog strategies since they provide an explicit representation of uncertainty [12, 6]. This enables the dialog manager to avoid and recover from recognition errors by sharing and shifting probability mass between multiple hypotheses of the current dialog state. However, scaling the dialog model to handle real-world problems remains a significant challenge for RL-based systems, given that the complexity of a POMDP grows with the number of user goals, and optimization quickly becomes intractable.

Several proposals have tried to reduce the size of the state space by using several state spaces [18], using approximate algorithms to overcome the intractability of exact algorithms [19], combining conventional dialog managers with a fully observable Markov decision process [20, 21, 6], using multiple POMDPs and selecting actions by means of handcrafted rules [22]. Research advances in RL for building spoken conversational interfaces have been reviewed and summarized in several surveys [23, 24, 25].

Corpus-based approaches to DM are usually based on the estimation of a statistical model from the sequences of the system and user dialog acts obtained from a set of training data. In [9], we proposed a corpus-based statistical dialog modeling methodology that employs a data-driven classification procedure to generate abstract representations of the system turns taking into account the previous history of the dialog. The benefits and flexibility of the proposed methodology were validated by developing statistical dialog managers for four spoken dialog systems of different complexity, designed for different languages and application domains (from transactional to problem-solving tasks).

Other interesting approaches for statistical dialog management model the conversational system by means of HMM [26], stochastic finite state transducers [27, 8], and Bayesian networks [28]. Also [29] proposed a different hybrid approach to dialog modeling in which n-best recognition hypotheses are weighted using a mixture of expert knowledge and data-driven measures by using an agenda and an example-based machine translation approach, respectively. Hybrid approaches to DM combine statistical and rule-based approaches to try to reduce the amount of dialog data required for parameter estimation and to allow system designers to directly incorporate their expert domain knowledge into the dialog management models [30].

## 2.2 Deep neural networks

Recently, encouraged by the positive results obtained in related areas, deep learning has started to be used for conversation [31]. Deep learning involves extracting patterns from data and classifying them by learning multiple layers of representation and abstraction [32, 33]. One of the benefits of deep learning is that the models and algorithms that are used in one application can be applied to a diverse range of other applications (e.g., computer vision, speech recognition, natural language understanding, audio processing, information retrieval, robotics, etc.).

The success of deep learning can be attributed to the availability of vast amounts of data, more powerful processors to process this data, modern Graphics Processing Units (GPUs), and new models for learning, in particular, a fast learning algorithm developed by Hinton and colleagues for learning deep belief networks [34].

A Deep Neural Network (DNN) differs from conventional neural networks in that it has multiple hidden layers between the input and output layers, providing greater learning capacity as well as the ability to model complex patterns of data.

Classic DNNs are fully connected feed-forward neural networks with at least 2 hidden layers of units between the input and output layer [35]. That means that only connections between units from a previous layer to the following layer are present. However, various architectures with different connections may be employed (e.g., convolutional neural networks [36], residual neural networks [37] or Recurrent Neural Networks (RNN) [38]). In particular, the backward connections are generalized in RNNs, and they are more challenging to learn. Recurrent neural networks are often used in natural language processing. An example of a feed-forward DNN architecture with 2 hidden layers with 3 units (neurons) per layer is depicted in Fig. 1.

In a fully connected feed-forward DNN, each neuron $j$ maps its total input $x_j$ from the previous layer to a real value $y_j$ (output of the neuron) that is passed to the next layer [39]. The total input of neuron $j$ is a linear function that can be expressed as follows:

$$x_j = b_j + \sum_{i}^{N} y_i w_{ij} ,\qquad(1)$$

where $b_j$ is a bias of neuron $j$. The variable $i$ corresponds to the index of the neuron in the previous layer, $N$ is the total number of neurons in the previous layer, $y_i$ stands for the output of neuron $i$, and $w_{ij}$ is a weight between neurons $i$ and $j$.

The mapping from total input $x_j$ to output $y_j$ is done using a nonlinear activation function. Within the scope of this work, 2 different activation functions are applied; sigmoid (2) and hyperbolic tangent (3);

$$y_j = \frac{1}{1 + e^{-x_j}} ,\qquad(2)$$

$$y_j = \frac{e^{2x_j} - 1}{e^{2x_j} + 1} .\qquad(3)$$

The activation function for the output layer of the DNN varies according to the final application. For multiclass classification, as used in this paper, a softmax activation function is employed. It is defined as:

$$y_j = \frac{e^{x_j}}{\sum_{k}^{K} e^{x_k}} ,\qquad(4)$$

where $x_j$ is the total input of neuron $j$, $K$ is the number of classifiable classes, and finally $k$ is the index over these classes. The output of the neuron is the probability of the given class.

The goal of DNN training is to update the weights of the DNN to represent the data more fittingly. The training is done using the backward propagation of errors (backpropagation) [40] of a loss (cost, error) function. The cost function expresses the difference between the actual output of the DNN and the target output (supervised) for each sample. Cross-entropy loss is a loss function often used with the softmax function. It is defined as follows:

$$E = -\sum_{j}^{J} d_j \log p_j \ ,\quad (5)$$

where $p_j$ is the probability of the output neuron $j$ and $d_j$ is the target probability. $J$ is the number of output neurons. The update function of the weights can then be expressed as:

$$\Delta w_{ij} = -\alpha \frac{\delta E}{\delta w_{ij}} \ ,\quad (6)$$

where $\alpha$ is the learning rate scaling the change of the original weight. $\delta E$ is the partial derivative of loss function with respect to a weight between neurons $i$ and $j$ ($w_{ij}$). The $-1$ assures the update is in the direction of the minimum of the cost function. Various strategies exist for updating the weights:

- after each sample,
- after $n$ amount of samples (mini-batches),
- after each epoch (a complete cycle through the training samples).

Easy use of the context of the inputs can be considered as one of the advantages of DNNs. This is usually done by concatenating the input feature vector with previous or even future samples. The process of DNN training thus remains unchanged. A more detailed formalism of DNN training can be found in [40].

### 2.3 Deep learning for conversational interfaces

Deep learning has led to substantial developments in speech-based interfaces [31, 41, 42]. DNNs are now used extensively in most commercially deployed Automatic Speech Recognition (ASR) systems [43, 44, 39, 35]. Several authors (e.g., [45, 46]) have described the deep learning approach to ASR using DNNs, and various studies have shown that DNNs outperform Gaussian Mixture Model – Hidden Markov Model (GMM-HMM) architecture in terms of increased speech recognition accuracy [39, 47]. These factors have been critical in enabling dramatic improvements in recognition accuracy due mainly to deeper and more precise acoustic modeling.

Similarly, deep learning methods have begun to outperform other machine-learning methods in spoken language understanding and NLP [48, 49, 50]. The deep NLP approach allows richer information to be captured compared to traditional vector-based approaches such as bag of words, in which structural relations between words and phrases are not captured. In recent work, a compositional vector grammar parser was used to jointly find syntactic structure and compositional semantic information by combining a probabilistic context-free grammar with an RNN [51]. Ravuri and Stolcke [52] were the first to propose the use of an RNN architecture for intent determination. These architectures have been extended to model both intent detection and slot filling in multiple domains [53, 54]. Deep learning has also been used successfully to perform many other NLP tasks, including part-of-speech tagging, chunking, semantic role labeling [50], and natural language generation for task-orientated dialog systems [55, 56].

The next area for development is conversational interaction. In task-oriented dialogs, compared to open-domain chat applications, the user pursues a predefined goal (e.g., book a train ticket). The success of the dialog depends on the quality of the policy that the dialog manager has in order to respond to the user inputs, which must in turn lead to the achievement of the user's goal.

Deep learning has been employed for dialog state tracking, that is, finding the current state of the dialog before making the decision of the next action. In [57], a DNN approach for dialog state tracking was proposed that showed promising results compared to using networks with fewer hidden layers. An enhanced word-based approach to dialog state tracking using recurrent neural networks that requires less feature engineering is described in [58]. The model is capable of generalizing to unseen dialog state hypotheses by using one RNN per slot, taking the current dialog turn (user input plus the last machine dialog act) as input, updating its internal memory and calculating an updated belief over the values for the slot.

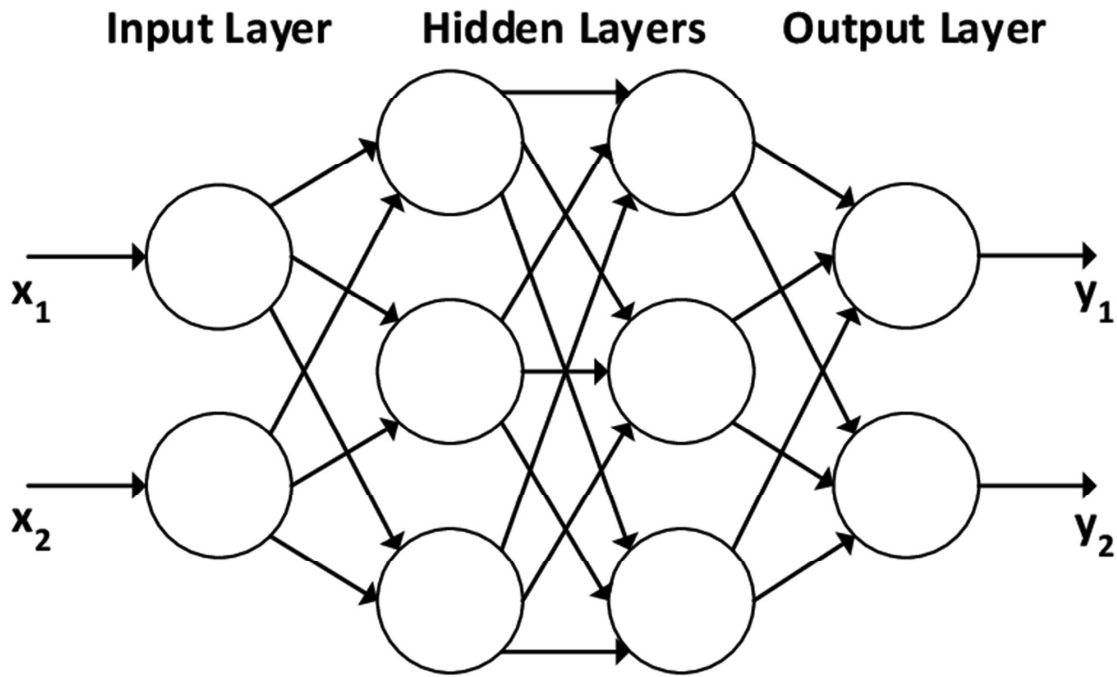

Figure 1: The architecture of a feed-forward DNN

Other proposals have also addressed the use of RNNs to build dialog trackers for open-domain dialog systems, avoiding the design of handcrafted domain-specific resources for semantic interpretation [59, 60]. The hierarchical training procedure proposed in [61, 62] first uses all the data available to train a very general belief tracking model that learns the most frequent and general dialog features present across the various domains. The general model is then specialized for each domain to learn domain-specific behavior while retaining the cross-domain dialog patterns previously learned. The practical application to a set of tasks showed that training using diverse dialog domains allowed the model to better capture general dialog dynamics applicable to different domains at once. Rastogi et al. [63] have also very recently proposed a multi-domain dialog state tracker to achieve effective and efficient domain adaptation.

Other uses of deep learning are coupled to RL, for example, interactive reinforcement learning methods have been applied to include human-provided rewards to learn the dialog policy [64, 65, 66, 67, 68], providing a good balance between supervised and reinforcement learning by enabling faster learning with users and non-expert crowd workers. In the proposals presented in [69, 64, 70], the agent receives action-specific feedback from the user and incorporates this feedback using recurrent neural networks for providing reward shaping and guiding the agent faster towards good dialog policies.

In summary, deep learning has a very recent trajectory within the area of RL dialog management and has been applied mainly in the preparatory steps, including dialog state tracking. The implications of deep learning for the selection of the next system action in real tasks are still open. Therefore, in this paper, we explore the application of DNNs for the development of a complete statistical dialog manager considering all the steps involved, i.e., the representation of the state of the dialog and the decision of the next system action considering different representations of the history of the dialog. We provide empirical results in several practical domains. Our approach employs a fully connected feed-forward deep neural network which is trained on labeled data from 3 dialog corpora. All the DNNs in our experiments were trained on a single GPU using the torch framework[1].

---

[1] `http://torch.ch/`

# 3 Data corpora

We have used three corpora for the evaluation of our proposal: DIHANA, Let's Go! and DSTC, the characteristics of which are summarized in Table 1.

Table 1: Main characteristics of the corpora

|                                                  | DIHANA | Let's Go! | DSTC    |
|--------------------------------------------------|--------|-----------|---------|
| Number of dialogs                                | 713    | 235       | 10,415  |
| Number of user turns                             | 4,002  | 4,222     | 122,025 |
| Average number of user turns per dialog          | 5.6    | 18.0      | 11.7    |
| Number of dimensions of the input feature vector | 19     | 11        | 13      |
| Number of possible system acts                   | 29     | 26        | 17      |

## 3.1 DIHANA corpus

A set of 900 dialogs was acquired in the DIHANA project[2], whose goal was the development of a dialog system providing railway information using spontaneous speech in Spanish [71]. Although this corpus was acquired using a Wizard of Oz technique (WOz), real speech recognition and understanding modules were used.

The dialogs in the corpus were labeled in terms of dialog acts. In the case of user turns, the dialog acts correspond to the classical frame representation of the meaning of the utterance. For the DIHANA task, eight concepts and ten attributes were defined. The eight concepts are divided into two groups:

1. *Task-dependent concepts*: they represent the concepts the user can ask for (*Hour*, *Price*, *Train-Type*, *Trip-Time*, and *Services*).

2. *Task-independent concepts*: they represent typical interactions in a dialog (*Affirmation*, *Negation*, and *Not-Understood*).

Three levels were defined for the labeling of the 51 system responses. The first level describes the general acts of any dialog, independently of the task. The second level represents the concepts and attributes involved. The third level represents the values of the attributes given in the turn. The following labels were defined for the first level: *Opening*, *Closing*, *Undefined*, *Not-Understood*, *Waiting*, *New-Query*, *Acceptance*, *Rejection*, *Question*, *Confirmation*, and *Answer*. The labels defined for the second and third level were the following: *Departure-Hour*, *Arrival-Hour*, *Price*, *Train-Type*, *Origin*, *Destination*, *Date*, *Order-Number*, *Number-Trains*, *Services*, *Class*, *Trip-Type*, *Trip-Time*, and *Nil*.

Figure 2 shows an example of the semantic interpretation of an input sentence.

| **Input sentence:**                                                                                 |
| [SPANISH] Sí, me gustaría conocer los horarios para esta tarde desde Valencia.                      |
| [ENGLISH] *Yes, I would like to know the timetables for this evening leaving from Valencia.*        |
| **Semantic interpretation:**                                                                        |
| (*Affirmation*)                                                                                     |
| (*Hour*)                                                                                            |
|    *Origin*: Valencia                                                                |
|    *Departure-Date*: Today                                                           |
|    *Departure-Hour*: Evening                                                         |

Figure 2: An example of the labeling of a user turn in the DIHANA corpus

## 3.2 Let's Go! corpus

Let's Go![3] is a spoken dialog system developed by the Carnegie Mellon University to provide bus schedule information in Pittsburgh at hours when the Port Authority phones are not answered by operators [72]. The information provided by the system covers a subset of 5 routes and 559 bus stops.

---

[2] http://universal.elra.info/product_info.php?products_id=1421
[3] http://www.speech.cs.cmu.edu/letsgo/letsgodata.html

The system has had many users since it was made available for the general public in 2005, so there is a substantial dataset that can be used to train a dialog model [73]. In addition, this large amount of data from spoken interactions has been acquired with real callers, rather than lab testers. For this research, we have used only a subset of real human-human dialogs that we used in previous work on other topics [74].

Each call to the system starts with a welcome message that prompts the user to make a request. Then, the system waits for the user's response and grabs concepts such as question type (e.g., *When is the next bus to X?*, *How can I go from X to Y?*) or departure and arrival times and places. To be successful, calls require three or four pieces of information from the user: a departure stop, a destination, a travel time, and, optionally, a bus route. Stops can be specified in one of three ways: the nearest intersection to the stop (e.g., Forbes [Avenue] at Murray [Avenue]), a neighborhood (e.g., Oakland), or a landmark or other point of interest (Pittsburgh International Airport, Waterworks Mall). The system explicitly prompts the user to provide the missing information to complete the query. Once the system has the required information to answer the user's query, it submits a query to the database, presents the results to the user, and prompts for a new query.

### 3.3 DSTC corpus

The Dialog State Tracking Challenge (DSTC)[4] is a regularly held community challenge which provides annotated data with dialog state information. The first challenge was held in 2013 [75]), while the most recent one (DSTC8) has been recently organized at the end of 2019. The main goals of these challenges remain the same: tracking the state of the dialog by estimating the user's goals as the dialog progresses and providing the community with a place to compare all the different state-of-the-art approaches to dialog state tracking.

Within the scope of this paper, we focus on the first challenge as it provides a substantial amount of data that can be fed to a deep neural network during the training phase. As with Let's Go!, the dialogs cover the bus timetable domain. In total, 10,415 dialogs with 122,025 dialog acts were extracted and used for DNN training and evaluation.

Each log of a human-computer dialog provides dialog acts as well as outputs from ASR and SLU units with probabilities to better track the user's intentions. However, in this paper, we focus rather on predicting the response of a system than the user's goals. For this reason, we had to relabel the data to better fit our needs. Each user's input was mapped into 12-dimensional feature vectors (DR and DCH). These inputs can be divided into 3 main groups: information (e.g., user provides where he wants to go), action (e.g., user confirms the action), and request (e.g., user asks about next bus). The complete list of possible inputs is listed in Table 2.

To react to the user's inputs, the system has 17 possible responses (shrunk from the original 30 by merging various error responses, requests, etc.). These can be grouped into four subsets: direct requests (e.g., ask the user where he wants to go), confirmations (e.g., ask if the target destination is correct), indirect confirmation + request, and system (hello, restart, error, please repeat, provide schedule).

Table 2: Possible user's inputs as represented in feature vector in our relabeling of the DSTC corpus

| Name | Type | Details |
| --- | --- | --- |
| route | information | user provided information about route |
| from | information | user provided information about origin |
| to | information | user provided information about destination |
| date | information | user provided information about date |
| time | information | user provided information about time |
| affirm | action | user provided affirmation to the system |
| negate | action | user provided negation to the system |
| nextbus | request | user requested next bus |
| prevbus | request | user requested previous bus |
| repeat | request | user asked for a repetition of last system act |
| bye | action | user ended the dialog |
| restart | request | user asked a restart of the dialog |

---

[4]https://www.microsoft.com/en-us/research/event/dialog-state-tracking-challenge/

# 4 Experimental setup

Our aim is to apply DNNs to learn dialog management policies, that is, to discern at each moment of the dialog what is the best next system dialog act.

In order to do this, we have checked empirically what are the implications of:

- The type of input features used for classification, that is, how the current state of the dialog is represented (see Section 4.1).
- The configuration of the hyper-parameters of DNN.
- The input context window size, i.e., how many previous turns in the dialog are considered for classification.
- The corpora used for training and testing (including their size, dimensionality and number of system outcomes as explained below.

The main differences between the corpora used are their size, that is, the amount of training data (number of dialog acts), the dimensionality of their inputs (number of possible user dialog acts), and number of possible outputs (possible system dialog acts), as shown in Table 1.

Regarding the size, while the DIHANA and Let's Go! corpora represent domains with a low amount of training data available, the DSTC corpus provides a larger amount of training data. With respect to dimensionality, the DIHANA corpus has the richest representation of the dialog state with a 19-dimensional feature vector.

As for the number of possible outputs, the DIHANA and Let's Go! corpora have a similar number of system acts from which the system response can be selected (29 and 26 system dialog acts), while the DSTC corpus only has 17 possibilities.

The metric used to compare and evaluate the results is Error Rate (ER):

$$ER[\%] = \frac{M}{N} * 100 \;, \tag{7}$$

where $M$ is the number of misclassified system dialog acts, and $N$ is the total number of system dialog acts. The final results are obtained by 5-fold cross-validation with 80% of the data used for training and 20% for testing.

## 4.1 Representation of the input

The choice of a feature extraction technique from the data is an integral part of a successful system. Within the scope of this work, two different types of feature sets are employed, which we have named: *dialog register* (DR) and *dialog changes* (DCH).

### 4.1.1 Dialog register

The objective of the dialog manager at each time $i$ is to find the best system response $A_i$. This selection is a local process for each time $i$ and takes into account the information shared by the user and the system during the previous history of the dialog.

To store the information provided by the user throughout the previous history of the dialog, we proposed in [9] the use of a data structure that we call dialog register (DR).

For the dialog manager to determine the next response, we assumed that the exact values of the fields in the $DR$ are not significant. They are important to access the databases and for constructing the output sentences of the system. However, the only information necessary to determine the next action by the system is the presence or absence of values in the fields. Therefore, the information we used from the $DR$ is a codification of this data in terms of only three values, $\{0, 1, 2\}$, for each field in the $DR$ according to the following criteria:

- **0**: The value of the field has not been given.
- **1**: The field contains a value with a confidence score that is higher than a given threshold (a value between 0 and 1). The confidence score is given during the recognition and understanding processes and can be increased by means of confirmation turns.
- **2**: The field contains a value with a confidence score that is lower than the given threshold.

The dialog registers keep the information of the whole dialog in each turn. Additionally, one of the features is a target class of the previous system dialog act. An example of a simple system providing information about train schedules could be: origin, destination, time, yes, and no. The first three features allow the system to monitor the dialog and act accordingly, the latter two provide the user with an option to react to the system output, either by confirming or denying the action.

### 4.1.2 Dialog changes

The dialog changes (DCH) structure contains the same slots and also includes information about the previous system dialog act. In this case, the values considered are:

- **0**: The value of the field has not changed in the current turn.
- **1**: The value of the field is provided with high confidence.
- **2**: The value of the field is provided with low confidence
- **4**: The value of a field that was previously recognized with low confidence is now provided with high confidence.

The main difference with respect to the DR is that it only keeps information about the changes in the current dialog turn. However, it has the advantage of keeping the order in which the pieces of data are provided, which is lost with DR. This approximation is similar to the approaches based on n-grams [76]. In our case, the value of N is defined in the experiments as the size of the context window. Fig. 3 shows an example of both structures.

| **Structure:** [Origin - Destination - Time - Yes - No] |
|---|
| **Input:** Hello! I'm going to Granada. [Score 0.95] <br> **DR:** [0 1 0 0 0] <br> **DCH:** [0 1 0 0 0] <br> *The destination is provided with high confidence.* |
| **Input:** I'm traveling from Madrid. [Score 0.15] <br> **DR:** [2 1 0 0 0] <br> **DCH:** [2 0 0 0 0] <br> *The origin is provided with low confidence, the DR does not lose track from the status of the destination, while for the DCH its value is 0 because it has not changed in the current turn.* |
| **Input:** Yes, from Madrid. Tomorrow morning. [Score 0.85] <br> **DR:** [1 1 1 1 0] <br> **DCH:** [4 0 1 1 0] <br> *The value for the origin changes from low to high confidence, the user has said "Yes" to confirm and provided a value for the time. The DR still keeps the information about the status of the destination, while for the DCH it has not changed.* |

Figure 3: An example of dialog registers and dialog changes

### 4.2 Description of the experiments

A summary of the experiments is depicted in Fig. 4. The first step was to extract the feature vectors (dialog registers and dialog changes) for all corpora. By employing these features, we trained baseline DNNs using the hyper-parameters that were most similar to the ones in our prior work with multilayer perceptrons [9]. Then, we conducted further experiments focused on balancing the output classes and selecting the best hyper-parameters for the DNNs comparing the feature vectors (DR and DCH) and different input context window sizes for all corpora.

## 5 Discussion of results

This section describes the experiments conducted and presents a discussion of their main results.

### 5.1 Baseline experiment

The baseline experiment utilized fully connected feed-forward deep neural networks. The hyper-parameters to train the DNN were set to be similar to those employed in our previous work with multilayer perceptrons [9]:

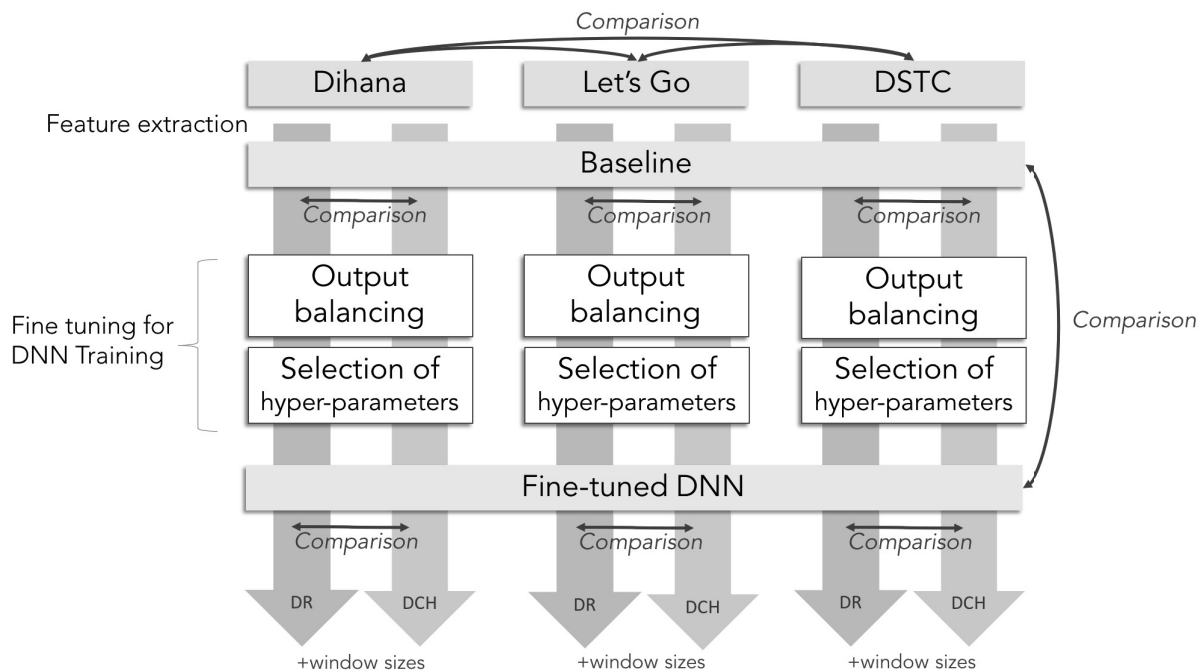

Figure 4: Summary of the experimental work

- 2 hidden layers,
- 100 and 10 neurons per hidden layer, respectively,
- sigmoid transfer function,
- learning rate $0.5$,
- mini-batches of size $8$,
- 50 epochs,
- global normalization of input features.

Regarding the context size, we experimented with a context window of sizes 0, 1, 2, and 3. In case of zero context, only the current turn is used for DNN training. With context size of $n$, the input feature is joint from the previous $n$ turns and the current dialog turn. In all cases, each turn is represented by using the feature vectors as presented in the previous section (either DR or DCH). An example is given in Fig. 5.

The results are presented in Tables 3, 4 and 5 for the DIHANA, Let's Go! and DSTC corpora, respectively.

Table 3: ER[%] for baseline experiment using DIHANA corpus

| Context | Dialog Registers | Dialog Changes |
|---|---|---|
| 0 | 17.8 | 40.0 |
| 1 | 22.2 | 38.3 |
| 2 | 27.2 | 39.6 |
| 3 | 30.8 | 40.0 |

For the DIHANA corpus (Table 3), the DR features significantly outperformed the DCH ones. The difference was 22.2% (17.8% versus 40%) without any context provided. This improvement may be due to the additional information stored in the DR structure as the DR keeps the information of the whole dialog while DCH only tracks the difference with respect to the last dialog turn. For DR, the best results were obtained by employing DNN trained on dialog turns with no additional context. This was most likely caused by an insufficient amount of training data (DNNs are usually trained on much larger sets of training data), and high-dimensionality of the input features (which increases the need for more training data even further). On the other hand, DCH benefited slightly (1.7%) from the context. This was due to the fact that DCH gain more information from the context window as the features only hold information about the current turn.

| Structure: [Origin - Destination - Time - Yes - No] |
|---|
| **Input:** Hello! I'm going to Granada. [Score 0.95]<br>**DR:** [0 1 0 0 0]<br>**Input feature vector:**<br>**Context 0:** [0 1 0 0 0]<br>**Context 1:** [0 0 0 0 0 0 1 0 0 0]<br>**Context 2:** [0 0 0 0 0 0 0 0 0 0 1 0 0 0]<br>*If there is no previous dialog turn, the feature is padded with zeros (no information given).* |
| **Input:** I'm traveling from Madrid. [Score 0.15]<br>**DR:** [2 1 0 0 0]<br>**Input feature vector:**<br>**Context 0:** [2 1 0 0 0]<br>**Context 1:** [0 1 0 0 0 2 1 0 0 0]<br>**Context 2:** [0 0 0 0 0 0 1 0 0 0 2 1 0 0 0]<br>*The previous turn is joint with the current dialog turn. There is still zero-padding in case of context size 2 as there is no information available.* |
| **Input:** Yes, from Madrid. Tomorrow morning [Score 0.85]<br>**DR:** [1 1 1 1 0]<br>**Input feature vector:**<br>**Context 0:** [1 1 1 1 0]<br>**Context 1:** [2 1 0 0 0 1 1 1 1 0]<br>**Context 2:** [0 1 0 0 0 2 1 0 0 0 1 1 1 1 0]<br>*Finally, no zero padding is employed as there is enough information in the dialog to provide full context for all cases.* |

Figure 5: An example of context usage

Table 4: ER[%] for baseline experiment using Let's Go! corpus

| Context | Dialog Registers | Dialog Changes |
|---:|---:|---:|
| 0 | 31.6 | 34.9 |
| 1 | 29.7 | 35.3 |
| 2 | 33.2 | 33.3 |
| 3 | 36.0 | 34.5 |

In previous work, we evaluated four different classifiers to develop dialog managers for the DIHANA task based on the DR structure defined for the task: naive Bayes classifier, n-gram based classifier, a classifier based on grammatical inference techniques and a classifier based on a multilayer perceptron (MLP) [9, 71]. The best results (ER[%]=25.6) were obtained using the MLP with one hidden layer of 32 units trained with the standard backpropagation algorithm and a value of LR equal to 0.3.

For the Let's Go! corpus (Table 4), the first thing to note is that the difference between the DR and DCH results was much smaller than for DIHANA. This was most likely caused by a lower dimensionality of the input feature vector and by the more consistent labeling of the training data in the case of DIHANA. Although the error rate was higher for Let's Go!, DR also outperformed DCH. The decrement in accuracy could be caused by multiple factors, including a higher amount of possible system acts for a similar number of training user turns. In addition, the low-dimensionality of the inputs made the context beneficial for both feature extraction techniques. The improvements were 1.1% and 1.6% for DR and DCH, respectively.

The Let's Go task is a common ground for experimentation and evaluation within the dialog system community [73, 77, 78, 79]. The initial version of the system provided complete dialogs with a 79% success rate with an average length of a dialog of 14 turns [80]. The version of the system presented in [78] provides a 77.64% success rate with an average number of turns of 11.47. With regard a version of the system developed by means of the DUDE development [79], the 62% of calls reached the stage of presenting results to the user. Of these calls, 61% gave fully correct information to the users, and 74% were correct with respect to the route information. The results of the evaluation of the different versions of the system that participated at the 2010 Spoken Dialog Challenge are compiled in [19]. The four systems described in this paper respectively provide success rates of $64.8 \pm 5.0$, $37.7 \pm 6.2$, $89.3 \pm 3.6$, and $74.7 \pm 4.8$.

Table 5: ER[%] for baseline experiment using DSTC corpus

| Context | Dialog Registers | Dialog Changes |
|---|---|---|
| 0 | 33.8 | 34.5 |
| 1 | 30.0 | 31.5 |
| 2 | 29.3 | 31.6 |
| 3 | 31.0 | 33.1 |

For the DSTC corpus (Table 5), the differences between the DR and DCH features were also smaller (2.3% for the best DNNs) than for DIHANA. This was probably caused again by the lower dimensionality of the corpus compared to DIHANA. The additional information stored in the DR also proved to be beneficial. Compared to Let's Go!, which is based on a similar domain, the bigger volume of data available for DSTC made the context even more important during the DNN training, so the best performing DNN was for a context window of size 2, higher than for the other corpora.

The baseline dialog tracker developed for the DSTC1 challenge maintains a single hypothesis for each slot (i.e. for each dialog state component). The best result for the joint goals defined at this challenge was an accuracy of 0.466 [81].

## 5.2 Balancing the outputs

The idea behind this experiment is to weight the DNN output probabilities by a priori probability of the output classes. This technique is often used in related areas (e.g., in speech recognition) to help with unbalanced training data.

The results are summarized in Table 6, Table 7 and Table 8 for DIHANA, Let's Go! and DSTC respectively. As can be observed, this technique worsened the results significantly. The increase in ER was significant for all the conducted experiments on all corpora. This happens because the number of occurrences of the least common classes (<10) is too low for the DNN to learn from it, and balancing the weight worsens the results as it basically boosts certain errors. Thus, the outputs were not balanced for the rest of the experiments.

Table 6: ER[%] for balancing the outputs experiment using DIHANA corpus

| Context | Dialog Registers | Dialog Changes |
|---|---|---|
| 0 | 31.4 | 55.2 |
| 1 | 34.7 | 52.9 |
| 2 | 39.1 | 51.9 |
| 3 | 40.3 | 51.3 |

Table 7: ER[%] for balancing the outputs experiment using Let's Go! corpus

| Context | Dialog Registers | Dialog Changes |
|---|---|---|
| 0 | 38.6 | 37.7 |
| 1 | 35.2 | 37.5 |
| 2 | 37.8 | 36.8 |
| 3 | 42.4 | 38.1 |

Table 8: ER[%] for balancing the outputs experiment using DSTC corpus

| Context | Dialog Registers | Dialog Changes |
|---|---|---|
| 0 | 46.5 | 47.7 |
| 1 | 41.0 | 43.5 |
| 2 | 40.9 | 45.0 |
| 3 | 43.6 | 45.4 |

## 5.3 Selection of the DNN hyper-parameters

In order to find the most optimal deep neural network configuration, we conducted a series of experiments focused on tuning the hyper-parameters of DNN. First, we fine-tuned the depth of the DNN (we experimented with shallow networks without a single hidden layer up to deeper networks with up to 6 hidden layers). The width of the hidden layer was also explored (more precisely, 8, 16, 32, 64, 128, 256, 512, and 1024 neurons per hidden layer configurations were tested). Next, we also worked with different transfer functions (sigmoid, hyperbolic tangent, and ReLU [82]) and tuned the associated learning rate. The use of dropout was also considered. Finally, all the DNNs were also evaluated after 5, 10, 15, 25, 50, 75, and 100 epochs. The final, best-performing, hyper-parameters (for all corpora) can be summarized as:

- 4 hidden layers,
- 64 neurons per layer,
- hyperbolic tangent transfer function,
- learning rate 0.01.

With the exception of these hyper-parameters, the rest remained the same as in the baseline setting. The increase in the depth and width of the deep neural network was possible due to the larger amount of dialog data available for deep neural network training. This also allowed us to model the target domain more accurately. The change in the activation function and the consecutive reduction in the learning rate resulted in faster DNN training. All the tuning of hyper-parameters yielded a decrease in error rate.

The results for DIHANA are shown in Table 9. The reduction in ER was 1.3% and 3.6% for the DR and DCH features, respectively, with no context employed. Additionally, for the dialog changes features, the optimization of the hyper-parameters eliminated the necessity of context. The improvement caused by the additional context information was once again prevented by an insufficient amount of training data and the high-dimensionality of input features.

Table 9: ER[%] for the fine-tuned DNN using the DIHANA corpus. The reduction in ER% with respect to the baseline is indicated in brackets

| Context | Dialog Registers | Dialog Changes |
|---|---|---|
| 0 | **16.5** [1.3] | **36.4** [3.6] |
| 1 | 18.7 [3.5] | 39.7 [-1.4] |
| 2 | 22.1 [5.1] | 39.9 [-0.3] |
| 3 | 22.5 [**8.3**] | 42.7 [-2.7] |

For Let's Go! the results are shown in Table 10. The reductions in ER were significant for both DR and DCH. Specifically, 5.8% and 10% reduction in ER of the best results for DR and DCH, respectively. Similarly to the baseline experiment, the use of a context window of size 1 yielded the best results.

Table 10: ER[%] for the fine-tuned DNN using the Let's Go! corpus. The reduction in ER% with respect to the baseline is indicated in brackets

| Context | Dialog Registers | Dialog Changes |
|---|---|---|
| 0 | 27.8 [3.8] | 27.2 [7.7] |
| 1 | **23.9** [5.8] | **25.3** [**10.0**] |
| 2 | 27.8 [5.4] | 28.0 [5.3] |
| 3 | 33.2 [2.8] | 32.4 [2.1] |

Table 11 shows the results for DSTC. A reduction of 3% in ER was achieved for all the experiments by tuning the hyper-parameters. Similarly to the baseline experiment, the context proved to be beneficial (4% and 3% reduction for DR and DCH, respectively). This was most likely caused by the larger amount of data available to train the deep neural network compared to the other corpora. With the use of the tuned hyper-parameters, the best size of the context window was 1 in contrast to the baseline experiment (2). Finally, the DR features also outperformed DCH for this domain.

The detailed analysis of the confusion matrices showed that, in some cases, the system did not detect that the user wanted to finish the dialog. A second problem was related to the introduction of data in the $DR$ with a high confidence value due to errors generated by the ASR that were not detected by the dialog manager. However, the evaluation confirms a good operation of the approach since the information is correctly given to the user in the majority of cases.

Table 11: ER[%] for the fine-tuned DNN using the DSTC corpus. The reduction in ER% with respect to the baseline is indicated in brackets

| Context | Dialog Registers | Dialog Changes |
|---------|------------------|----------------|
| 0 | 30.2 [3.6] | 31.4 [3.1] |
| 1 | **26.0 [4.0]** | **28.1** [3.4] |
| 2 | 26.5 [2.8] | 28.4 [3.2] |
| 3 | 27.3 [3.7] | 29.4 [**3.7**] |

## 6 Comparison of results

This section summarizes the results achieved over all 3 corpora.

### 6.1 Comparison of feature extraction techniques

The dialog register features outperformed the dialog changes approach on all corpora. The difference was higher for DIHANA, where it was 19.9% for the best-achieved results. For the other corpora, the difference was 1.4 and 2.1 for the Let's Go! and DSTC corpora, respectively%. The significant difference on the DIHANA corpus was most likely caused by the high dimensionality of the input feature vectors (the dialog registers can provide more information than for other corpora).

### 6.2 The effect of the context

The results confirmed that, as expected, the importance of context grew with the amount of training data. However, another observation became clear during the evaluation, the dimensionality of input features mattered as well. The use of context did not reduce ER on the DIHANA corpus due to the low amount of training data and the high dimensionality of inputs compared to, for instance, DSTC. However, a size of 1 for the context window helped with the other corpora (the reduction was around 4% ER for dialog registers). Both used lower-dimensional input feature vectors. Additionally, for DSTC, the large amount of training data made the results with wider context window sizes more competitive. These results can be compared in Figure 6.

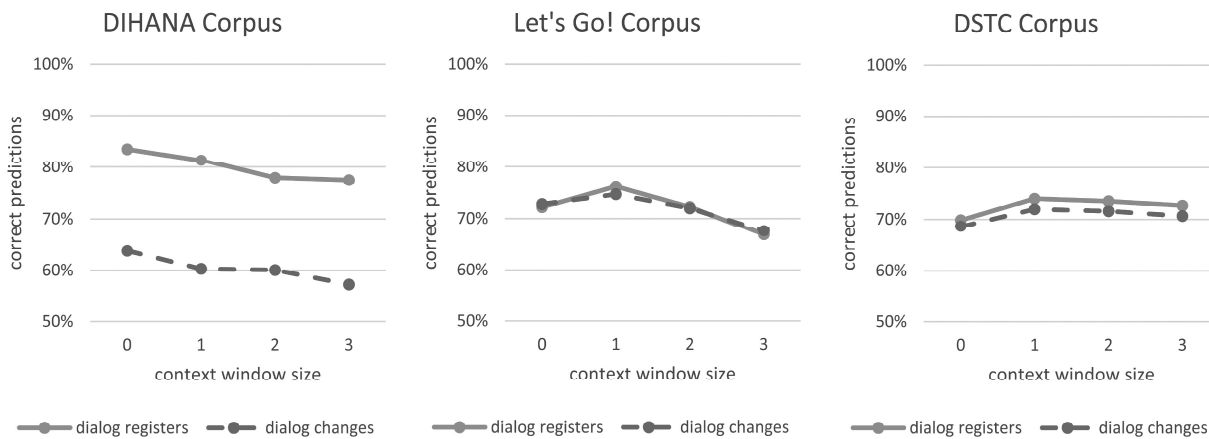

Figure 6: Correctly predicted system dialog acts based on context window size per different corpora.

### 6.3 The difficulty of the domain

The DIHANA corpus obtained the highest accuracy rates. This was most likely caused by the quality of the corpus itself. The high dimensionality of the inputs also provided more information to the DNN training. The results for the other corpora were similar even though the Let's Go! corpus was slightly easier to predict. The worse performance could be caused by the low amount of training data for the Let's Go! corpus (only 235 dialogs) and by the wide variety of dialogs in the DSTC data. The DSTC corpus also consisted of a high number of failed dialogs (that could be cleaned in further attempts to reduce ER).

# 7 Conclusions and future work

Deep learning technologies have recently helped to improve different aspects of human-machine communication, including speech recognition and understanding. However, their application to improve dialog management is starting to be studied and is usually restricted to specific phases, such as tracking the current dialog state.

The objective of dialog management is to find the most adequate system response (system dialog act) to certain user input (current user dialog act) given the knowledge available about the context in which the input was produced. When training a deep neural network to perform this task, there are several aspects that may play an important role, including the training data and how it is represented, the inclusion of context and the configuration of the neural networks.

This paper presents a series of experiments to obtain empirical evidence of the relevant aspects that can help to improve the accuracy of the classification process. In order to do so, we have used three corpora corresponding to different application domains and with a different size and dimensionality, and we have compared different ways of creating the feature vectors, including knowledge about the history of the dialog, introducing context and configuring the networks.

Our results show that the DNN works best in all conditions and corpora with the dialog register instead of the dialog changes structure. The codification of the information and the definition of this data structure which takes into account the data supplied by the user throughout the dialog makes possible to isolate task-dependent knowledge and apply our proposal to real practical domains. This shows that encoding the history of the dialog within the feature vector obtains better classification results. The DNN is capable of extracting this stored information to model and predict the dialog better.

Additionally, the smaller number of potential dialog register combinations (e.g., 3 possible values for each slot) requires a significantly lower amount of training data than dialog changes (e.g., 9 possible values) to train the network properly. To make full use of dialog changes, a large corpus would be required (i.e., the training corpus should be considered for feature selection). For instance, the Let's Go task considers up to 455 different values for the task-dependent user dialog acts. Our codification of the features using only 3 values $(0, 1, 2)$ makes it possible to reduce the dimensionality of the problem making the dialog manager more efficient and scalable, and suitable for complex application domains. Finally, higher dimensionality of the feature vector has a positive impact, compensating for a higher number of possible outputs.

In addition, context windows have a positive effect on the dialog changes input structure as they introduce information about the previous turns, but still its effect is less positive than using the dialog register. The effect of the window is more significant in larger corpora. Also, the fine-tuning of the DNN hyper-parameters provided an improvement over the baseline configuration. These results have relevant implications for how dialog corpora can be considered for DNN-based dialog management in terms of feature extraction, input representation, context consideration and DNN tuning.

For future work, we plan to perform a detailed analysis of the confusion matrices obtained, as not every misclassified dialog act is actually negative for the dialog, rather it can be a coherent alternative that does not match the response selected in the initial dialog. This would allow us to compute the relevance measures in more representative form. The corpora used have the standard size in the dialog management application domain (the DSTC corpus can even be considered the largest that is openly available to the community). We will also explore the benefits that can be derived from including additional samples by means of user simulation techniques.

## Acknowledgements

This work was supported by the Student Grant Scheme 2020 of the Technical University in Liberec, the European Union's Horizon 2020 research and innovation programme under grant agreement No 823907 (MENHIR project: https://menhir-project.eu), and by the Spanish project TEC2017-84593-C2-1-R.